# Kuantum Çağında Makine Öğrenimi: Kuantum ve Klasik Destek Vektör Makinelerinin Karşı Karşıya Gelmesi

## Machine Learning in the Quantum Age: Quantum vs. Classical Support Vector Machines

Davut Emre TAŞAR[1]
*Dokuz Eylül Üniversitesi*
*İktisadi ve İdari Bilimler Fakültesi*
*İzmir, Türkiye*
davutemre.tasar@ogr.deu.edu.tr
ORCID:0000-0002-7788-0478

Kutan KORUYAN[1]
*Dokuz Eylül Üniversitesi*
*İktisadi ve İdari Bilimler Fakültesi*
*İzmir, Türkiye*
kutan.koruyan@deu.edu.tr
ORCID: 0000-0002-3115-5676

Ceren ÖCAL TAŞAR[1]
*Bağımsız Araştırmacı*

*İzmir, Türkiye*
ceren.ocaltasar@gmail.com
ORCID: 0000-0002-0652-7386

**Öz**

*Bu çalışma, klasik ve kuantum hesaplama paradigmalarındaki makine öğrenimi algoritmalarının performansını karşılaştırmayı amaçlamaktadır. Özellikle, Destek Vektör Makineleri (SVM) üzerinde durarak, klasik SVM ile kuantum donanımı üzerinde çalıştırılan Kuantum Destek Vektör Makineleri (QSVM)'nin Iris veri seti üzerindeki sınıflandırma başarısını değerlendirmekteyiz. Kullanılan metodoloji, Qiskit kütüphanesi ile gerçekleştirilen kapsamlı deneyler serisini ve hiperparametre optimizasyonunu içermektedir. Elde edilen sonuçlar, belirli durumlarda QSVM'lerin klasik SVM'lerle rekabet edebilecek düzeyde doğruluk sağladığını, fakat çalışma sürelerinin şu an için daha uzun olduğunu göstermektedir. Ayrıca, kuantum hesaplama kapasitesinin ve paralellik derecesinin arttırılmasının, kuantum makine öğrenimi algoritmalarının performansını önemli ölçüde iyileştirebileceğini belirtmekteyiz. Bu çalışma, kuantum çağında makine öğrenimi uygulamalarının mevcut durumu ve gelecekteki potansiyeli hakkında değerli içgörüler sunmaktadır. Colab: https://t.ly/QKuz0*

**Anahtar sözcükler:** Kuantum Hesaplama, Kuantum Makine Öğrenimi, Kuantum Destek Vektör Makineleri (QSVM), Klasik Destek Vektör Makineleri (SVM), Iris Veri Seti, Qiskit Kütüphanesi, Hiperparametre Optimizasyonu

**Abstract**

*This work endeavors to juxtapose the efficacy of machine learning algorithms within classical and quantum computational paradigms. Particularly, by emphasizing on Support Vector Machines (SVM), we scrutinize the classification prowess of classical SVM and Quantum Support Vector Machines (QSVM) operational on quantum hardware over the Iris dataset. The methodology embraced encapsulates an extensive array of experiments orchestrated through the Qiskit library, alongside hyperparameter optimization. The findings unveil that in particular scenarios, QSVMs extend a level of accuracy that can vie with classical SVMs, albeit the execution times are presently protracted. Moreover, we underscore that augmenting quantum computational capacity and the magnitude of parallelism can markedly ameliorate the performance of quantum machine learning algorithms. This inquiry furnishes invaluable insights regarding the extant scenario and future potentiality of machine learning applications in the quantum epoch. Colab: https://t.ly/QKuz0*

# 1. Giriş

Son yıllarda, kuantum hesaplamanın yükselişi ve gelişen teknolojik kapasiteler, hem akademik hem de endüstriyel araştırmacılar için makine öğrenimi (ML) algoritmalarının potansiyelini keşfetme olanağı sunmuştur[1]. Kuantum hesaplama, süperpozisyon ve örtüşme gibi kuantum mekaniği prensiplerini kullanarak, klasik bilgisayarların ulaşamayacağı hesaplama kapasitelerine erişim sağlar[2]. Bu durum, ML algoritmalarının performansını ve ölçeklenebilirliğini artırma potansiyelini beraberinde getirir. Bu çalışma, klasik ve kuantum ML algoritmalarının performansını, özellikle Destek Vektör Makineleri (SVM) ve Kuantum Destek Vektör Makineleri (QSVM) üzerinden karşılaştırarak, mevcut durumu ve gelecek potansiyellerini değerlendirmeyi amaçlar. Qiskit ve Scikit-learn kütüphaneleri kullanılarak gerçekleştirilen deneylerle, klasik ve kuantum ML algoritmalarının, Iris veri seti üzerindeki sınıflandırma başarımı ve QSVM hiperparametre optimizasyonu incelenmiştir.

## 1.1 Makine Öğrenmesi

Makine öğrenmesi (ML) terimi, ilk olarak 1959 yılında Arthur Samuel tarafından ortaya atılmış ve bu alandaki erken çalışmaları temsil etmiştir[3]. Samuel, makine öğrenmesini "bilgisayarların açıkça programlanmadan öğrenme yeteneği" olarak tanımlamıştır[3]. Zamanla, ML algoritmaları ve modelleri, geniş bir uygulama yelpazesi bulmuş ve önemli bir araştırma alanı haline gelmiştir. Özellikle, Geoffrey Hinton, Yann LeCun ve Yoshua Bengio'nun derin öğrenme üzerine yaptığı çalışmalar[4][5], ML'nin bu alandaki etkisini göstermektedir. En çok atıf alan makine öğrenmesi çalışmalarından biri olan "Understanding Machine Learning: From Theory to Algorithms" başlıklı makale, ML'nin teorik ve algoritmik temellerini anlatmakta ve ML'nin geniş bir uygulama yelpazesi olan veri madenciliği, otomatik metin tanıma, görüntü işleme ve biyoinformatik gibi alanlarda nasıl kullanılabileceğini tartışmaktadır[6].

## 1.2 Quantum Makine Öğrenmesi

Quantum Makine Öğrenmesi (QML) kuantum bilgisayarlarının ilkelerini klasik makine öğrenimi teknikleri ile, veri analizi için yeni bir yaklaşım sunar. Bu alan, Feynman'ın 1981 yılında klasik bilgisayarlar için mümkün olmayan kuantum sistemlerine alternatif olması amacıyla kuantum bilgisayarlarını kullanma önerisine dayanır[7]. QML'nin amacı, kuantum algoritmalarını ve kuantum hesaplama modellerini kullanarak makine öğrenmesi algoritmalarını geliştirmektir. Kuantum algoritmalar, klasik algoritmaların aksine, süperpozisyon ve örtüşme gibi kuantum mekaniği ilkelerini kullanarak büyük veri setlerini hızlı bir şekilde işleyebilir. Kuantum Makine Öğrenimi'nin temel bir algoritması, Aram Harrow, Avinatan Hassidim ve Seth Lloyd tarafından lineer denklem sistemlerini çözmek için önerilen bir kuantum algoritmasıdır[8]. Kuantum Makine Öğrenimi, kuantum bilgisayarların klasik modellerden daha sağlıklı bir maliyetle daha geniş ölçekli makine öğrenmesi modelleri sağlama yeteneklerini kullanmayı amaçlar[9]. Bu alan hızla büyümekte ve birçok farklı algoritmayı ve uygulamayı içermektedir, 2017 sonrasında yayınlanan makalelerde çeşitli algoritmalar ve uygulamalar tanımlanmış, analiz edilmiş ve uygulamaları ele alınmıştır[10]. Kuantum Makine Öğrenimi, kuantum ve klasik bilgisayarların birleşiminden elde edilen hızlı veri analizi yetenekleri ile klasik makine öğrenimi ve kuantum bilgisayar bilimi arasında bir köprü oluşturur[11][12].

# 2. Veri, Yöntem ve Deney

## 2.1 Veri

Bu çalışma için, klasik ve kuantum makine öğrenmesi algoritmalarının performansını karşılaştırmak amacıyla Iris veri seti kullanılmıştır. Iris veri seti, İngiliz istatistikçi ve biyolog Ronald Fisher tarafından 1936 yılında tanıtılmış ve çok değişkenli bir veri seti olarak kullanılmıştır. Bu veri seti, üç farklı Iris türünün (Iris setosa, Iris virginica ve Iris versicolor) taç yaprak ve çanak yaprak boyutlarını içerir. Her bir türden 50 kayıt olmak üzere toplamda 150 kayıt bulunur. Veri setinin her bir satırı bir çiçeği, sütunlar ise çanak yaprak uzunluğu, çanak yaprak genişliği, taç yaprak uzunluğu ve taç yaprak genişliği özelliklerini temsil eder[13]. Çizelge 1'de veri setinin temsili gösterimi bulunmaktadır. Ölçüler cm cinsinden ifade edilmektedir.

**Çizelge-1: Iris Veri Setinin Temsili Gösterimi**

| Tür | Çanak Yaprak Uzunluğu | Çanak Yaprak Genişliği | Taç Yaprak Uzunluğu | Taç Yaprak Genişliği |
|---|---|---|---|---|
| Iris-setosa | 5.1 | 3.5 | 1.4 | 0.2 |
| Iris-versicolor | 7.0 | 3.2 | 4.7 | 1.4 |
| Iris-virginica | 6.3 | 3.3 | 6.0 | 2.5 |

Veri seti, Scikit-learn kütüphanesinin datasets modülü kullanılarak yüklenmiştir. Iris veri seti, load_iris fonksiyonu aracılığıyla iris değişkenine atanmış, ve bağımsız değişkenler (X) ile bağımlı değişkenler (y) ayrı ayrı değişkenlere atanmıştır.

## 2.2 Yöntem

### 2.2.1 Destek Vektör Makineleri

Destek Vektör Makineleri (SVM), sınıflandırma ve regresyon analizleri gibi çeşitli görevler için kullanılan güçlü ve esnek bir makine öğrenimi modelidir. SVM, 1995 yılında Cortes ve Vapnik tarafından geliştirilmiştir. Bu model, veri noktalarını ayıran bir marjin maksimize eden bir karar hiperdüzlemi oluşturmayı amaçlar. Maksimum marjin, modelin genelleştirme yeteneğini artırır ve overfitting riskini azaltır.

SVM, doğrusal ve doğrusal olmayan sınıflandırma görevlerinde etkili bir şekilde çalışabilir. Doğrusal olmayan sınıflandırma için, SVM, kernel trick adı verilen bir teknik kullanır. Bu teknik, veriyi daha yüksek boyutlu bir uzaya dönüştürerek doğrusal olarak ayrılabilir hale getirir[14].

Destek vektörleri, karar hiperdüzleminin oluşturulmasında kritik öneme sahiptir. Bu vektörler, karar hiperdüzlemine en yakın veri noktalarıdır ve marjinin hesaplanmasında esas alınır. Marjin, destek vektörlerine olan en kısa mesafeyi temsil eder[15]. Şekil-1'de, doğrusal SVM sınıflandırmasının iki farklı senaryosu temsil edilmektedir. İki belirleyici (X1 ve X2) temelinde, iki farklı sınıf mavi daireler ve kırmızı üçgenler ile gösterilmiştir. Her bir grafikte, sınıflandırma hiperdüzlemi (H) ve maksimal marjin (M) belirgin bir şekilde gösterilmiştir. Destek vektörleri ise açık daireler ve üçgenler ile belirtilmiştir.

Sol taraftaki görselde: Karar sınırı doğrusal olarak ayrılabilir ve tüm gözlemler mükemmel bir şekilde sınıflandırılmıştır. Bu durum, verilerin doğrusal olarak ayrılabilir olduğu ve karar hiperdüzlemi ile iki sınıfı tamamen doğru bir şekilde ayırabileceği senaryoyu temsil eder.

Sağ taraftaki görselde: Karar sınırı doğrusal olmasına rağmen, tüm gözlemleri mükemmel bir şekilde sınıflandıramamaktadır. Bu durum, yumuşak marjinlerin kullanıldığı ve karar hiperdüzlemi ile bazı gözlemlerin yanlış sınıflandırıldığı bir senaryoyu temsil eder. Yumuşak marjin kullanımı, SVM'nin doğrusal olmayan sınıflandırma problemleri üzerinde daha esnek ve tolere bir performans sergilemesini sağlar.

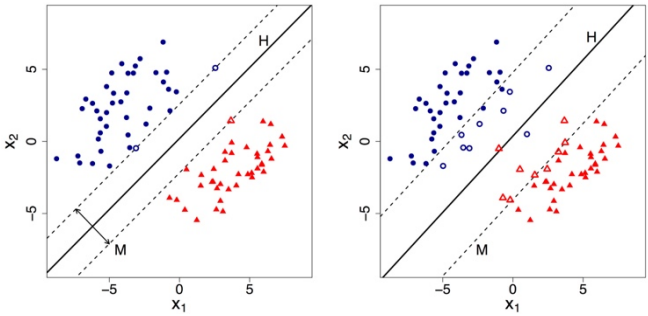

**Şekil-1**: X1 ve X2 Değişkenlerine dayalı doğrusal destek vektör makinası sınıflandırma örneği[16]

SVM modelinin başarısını artırmak ve daha iyi sınıflandırma sonuçları elde etmek için hiperparametre optimizasyonu önemli bir adımdır. SVM'nin temel hiperparametreleri arasında C, kernel ve gamma bulunur. C parametresi, yanlış sınıflandırma durumlarına karşı toleransı kontrol eder. Kernel parametresi, kullanılacak kernel fonksiyonunu belirtir, ve gamma parametresi, RBF kernel fonksiyonunun genişliğini kontrol eder[17][18][19].

SVM, çeşitli gerçek dünya uygulamalarında başarı ile kullanılmıştır. Özellikle, metin ve görüntü sınıflandırma, doğal dil işleme çıktılarının tabular data olarak sınıflandırılması, yüz tanıma, biyoinformatik ve medikal görüntü analizi gibi alanlarda geniş bir uygulama yelpazesi bulunmaktadır[20][21].

SVM, Qiskit kütüphanesinin bir parçası olarak quantum destek vektör makineleri (QSVM) ile birlikte, bu çalışmada kullanılan metodolojilerin temelini oluşturmaktadır. Iris veri seti üzerindeki deneyler, SVM ve QSVM modellerinin sınıflandırma başarısını karşılaştırmayı ve kuantum makine öğrenimi algoritmalarının performansını değerlendirmeyi amaçlamaktadır.

### 2.2.2. Quantum Destek Vektör Makineleri

Quantum Destek Vektör Makineleri (QSVM), kuantum hesaplama prensiplerini temel alan ve klasik Destek Vektör Makineleri'nin (SVM) bir uzantısı olarak geliştirilen yenilikçi bir makine öğrenimi modelidir[22]. QSVM, klasik SVM'nin doğrusal ve doğrusal olmayan sınıflandırma yeteneklerini, kuantum mekaniksel süreçlerin avantajları ile birleştirir. Kuantum hesaplaması, süperpozisyon ve örtüşme gibi kuantum mekaniği ilkelerini kullanarak, klasik hesaplamaların ulaşabileceği sınırları aşmayı amaçlar. Bu bağlamda, QSVM, kuantum hızlandırıcıları kullanarak sınıflandırma problemlerini daha hızlı ve etkili bir şekilde çözmeyi hedefler.

QSVM'nin temelinde, kuantum kernel metodları bulunur. Klasik kernel metodları, verileri yüksek boyutlu bir uzaya dönüştürerek doğrusal olarak ayrılabilir hale getirirken, kuantum kernel metodları, kuantum süperpozisyon ve örtüşme prensiplerini kullanarak bu dönüşümü gerçekleştirir. Bu, QSVM'nin karmaşık ve yüksek boyutlu veri setlerini daha etkili bir şekilde işlemesine olanak tanır[23].

QSVM algoritmalarının uygulanması, kuantum donanım ve yazılım platformlarına bağlıdır. Bu çalışmada, IBM'in Qiskit kütüphanesi kullanılarak QSVM modelleri kurulmuş ve eğitilmiştir. Qiskit, kuantum algoritmalarının geliştirilmesi ve simülasyonu için güçlü bir araç seti sunar[24]. QSVM'nin performansı, klasik SVM modeli ile karşılaştırılarak Iris veri seti üzerinde değerlendirilmiştir.

Hyperparametre optimizasyonu, QSVM modellerinin performansını artırmak için kritik bir öneme sahiptir. Qiskit kütüphanesinin sağladığı araçlar, C, kernel ve gamma gibi hyperparametrelerin optimizasyonunu sağlar, bu da modelin sınıflandırma doğruluğunu önemli ölçüde artırabilir[25].

Deney bölümünde, QSVM'nin Qiskit kütüphanesi kullanılarak nasıl implemente edildiği, ve QSVM'nin klasik SVM modeli ile karşılaştırıldığında sınıflandırma performansının nasıl farklılık gösterdiği açıklanacaktır. Ayrıca, kuantum kernel metodlarının klasik kernel metodlarına göre avantajları ve sınırlamaları tartışılacaktır. QSVM'nin kuantum hesaplama kapasitesini ve paralellik derecesini artırarak sınıflandırma performansını nasıl iyileştirebileceği üzerinde durulacaktır.

QSVM, kuantum makine öğrenimi algoritmalarının gelecekteki potansiyelini anlamak ve kuantum hesaplamanın makine öğrenimi uygulamalarındaki rolünü değerlendirmek için önemli bir örnektir. Iris veri seti üzerinde gerçekleştirilen deneyler, QSVM'nin klasik SVM modelleri ile rekabet edebilecek düzeyde performans sunabileceğini göstermektedir.

### 2.3. Deney

Bu bölümde, makalede bahsedilen metodolojileri kullanarak gerçekleştirilen deneylerin detaylarına yer verilmiştir. Deneyler Python programlama dilinde Qiskit ve Scikit-learn kütüphaneleri kullanılarak yürütülmüştür. İlk olarak, Iris veri seti Scikit-learn kütüphanesinin datasets modülü ile yüklenmiş ve train_test_split fonksiyonu aracılığıyla eğitim ve test setlerine ayrılmıştır. Daha sonra, StandardScaler kullanılarak özellikler ölçeklendirilmiştir.

Klasik Destek Vektör Makineleri (SVM) modeli, Radial Basis Function (RBF) kerneli ile oluşturulmuştur[26]. RBF kerneli, iki örnek arasındaki benzerliği ölçen bir fonksiyon olarak işlev görür ve bu örnekte Gaussian RBF olarak kullanılmıştır. Olasılıkları hesaplamak amacıyla probability=True parametresi belirlenmiştir, böylece model sınıflandırma olasılıklarını hesaplayabilir. SVC sınıfı kullanılarak model kurulmuş ve eğitilmiştir. Modelin eğitim süresi, time modülü aracılığıyla kaydedilmiştir ve sınıflandırma raporu classification_report fonksiyonu ile elde edilmiştir.

Kuantum Destek Vektör Makineleri (QSVM) için, Qiskit kütüphanesinin farklı özellik haritaları ve entanglement tipleri kullanılarak bir dizi deney gerçekleştirilmiştir. Her bir deneyde, ZFeatureMap, ZZFeatureMap ve PauliFeatureMap sınıfları kullanılmıştır. ZFeatureMap, veri örneklerini kuantum durumlarına haritalayan bir özellik haritasıdır. ZZFeatureMap, kuantum durumları arasındaki çift karşılıklı etkileşimi haritalayan bir özellik haritasıdır, ve PauliFeatureMap, Pauli matrislerini kullanarak özellikleri haritalayan bir özellik haritasıdır[24][27]. Linear, circular ve full entanglement tipleri kullanılmıştır. Linear entanglement tipinde, iki qubit arasında sadece bitişik qubitler birbirine bağlanır. Circular entanglement tipinde, her qubit bir sonraki ile ve son qubit de ilk qubit ile bağlanır. Full entanglement tipinde ise, her qubit her diğer qubit ile bağlanır[28]. QSVC modeli, QuantumKernel sınıfı ve Qiskit'in qasm_simulator backend'i kullanılarak kurulmuş ve eğitilmiştir. QuantumKernel sınıfı, kuantum mekaniği ilkelerini kullanarak bir kernel hesaplar. qasm_simulator backend'i[24], bir kuantum devresinin simülasyonunu sağlar. Her bir modelin eğitim süresi kaydedilmiş ve sınıflandırma raporu elde edilmiştir[24].

Son olarak, en iyi sonuç elde edilen modele, hiperparametre optimizasyonu için bir grid search uygulanmış ve farklı reps, C ve gamma değerleri ile QSVM modelleri eğitilmiştir. Reps parametresi, özellik haritasındaki devre tekrar sayısını belirtir. C parametresi, hata terimi üzerindeki düzenleme parametresini belirtir. Gamma parametresi ise, RBF kernelinin genişliğini belirtir. Her bir modelin doğruluk, F1-Skoru ve çalışma süresi kaydedilmiş ve en iyi model, en yüksek F1-Skoruna göre seçilmiştir.

## 3. Sonuç

Bu çalışmada, hem klasik hem de kuantum destek vektör makineleri (SVM ve QSVM) kullanılarak sınıflandırma deneyleri gerçekleştirilmiştir. Klasik SVM, Gaussian RBF kerneli kullanılarak uygulanmış ve mükemmel bir doğruluk oranı elde etmiştir. Bununla birlikte, kuantum algoritmalarının potansiyelini değerlendirmek amacıyla, farklı özellik haritaları ve entanglement stratejileri kullanarak bir dizi QSVM deneyi Quantum Simulator kullanarak gerçekleştirilmiştir.

İlk aşamada, ZFeatureMap, ZZFeatureMap ve PauliFeatureMap özellik haritaları üzerinde çeşitli entanglement stratejileri test edilmiştir. Yapılan analizler, ZFeatureMap'in lineer ve dairesel entanglement stratejileri ile en yüksek doğruluk oranlarını gösterdiğini ortaya koymuştur. Bu sonuçlar, belirli özellik haritalarının ve entanglement stratejilerinin QSVM'nin performansı üzerinde belirleyici bir etkiye sahip olduğunu göstermektedir. Bununla birlikte, QSVM'nin simülasyon süresi, klasik SVM'ye kıyasla önemli ölçüde daha uzundur. Bu, özellikle yüksek hesaplama gereksinimleri nedeniyle kuantum simülasyonlarının doğası gereğidir. Çizelge-2'de ilk deneyin sonuçları karşılaştırılmıştır.

**Çizelge-2: Klasik SVC ile Quantum SVC 1. Kademe Yöntem Arama Karşılaştırması**

| Metodoloji | Özellik Haritası (Feature Map) | Entanglement | F1-Score (Macro) | Simulatör Çalışma Süresi (saniye) |
|---|---|---|---|---|
| **Klasik SVC** | - | - | **1.00** | **0.002** |
| Quantum SVC | Z | Linear | **0.96** | 58.418 |
| **Quantum SVC** | Z | **Circular** | **0.96** | **38.237** |
| Quantum SVC | Z | Full | 0.96 | 34.960 |
| Quantum SVC | ZZ | Linear | 0.45 | 59.371 |
| Quantum SVC | ZZ | Circular | 0.43 | 54.979 |
| Quantum SVC | ZZ | Full | 0.43 | 65.219 |
| Quantum SVC | Pauli | Linear | 0.76 | 52.692 |
| Quantum SVC | Pauli | Circular | 0.74 | 52.906 |
| Quantum SVC | Pauli | Full | 0.72 | 52.922 |

İkinci aşamada, en iyi performans gösteren ZFeatureMap özelliği ile hiperparametre optimizasyonu uygulanmıştır. Çizelge-3'te farklı "reps", "C" ve "gamma" değerleri üzerinde yapılan grid search'in sonuçlarını göstermektedir. En iyi sonuç, "reps=1, C=10, gamma=scale" parametreleri ile elde edilmiş ve bu model, klasik SVM ile karşılaştırılabilir mükemmel bir doğruluk oranı sunmuştur.

Bu çalışma, kuantum destek vektör makinelerinin (QSVM) potansiyelini ve klasik SVM yaklaşımlarına kıyasla performansını detaylı bir şekilde ortaya koymaktadır. Ancak, uygulamada karşılaşılan zorluklar ve sınırlamalar, kuantum hesaplamanın pratik uygulamalarını genişletme yolunda atılması gereken adımları da vurgulamaktadır.

Bir engel, QSVM'nin yüksek hesaplama süreleriyle ilişkilidir. Bu, özellikle kuantum durumlarının klasik bilgisayarlarda simülasyonu söz konusu olduğunda belirgindir, çünkü bu durumlar genellikle büyük miktarda hesaplama kaynağı gerektirir. Ayrıca, kullanılan entanglement stratejileri ve özellik haritalarının içsel karmaşıklığı, simülasyon sürelerinin artmasına neden olabilir, özellikle de çok sayıda qubitin işin içine girdiği durumlarda. Simülasyon altyapısının kendine özgü kısıtlamaları da bu süreçte bir diğer değişkeni oluşturur.

Ancak, teknolojik ilerlemeler ve kuantum donanımının evrimi, bu hesaplama sürelerini radikal bir şekilde azaltma potansiyeline sahiptir. Daha verimli kuantum simülasyon teknikleri ve algoritmalarının yanı sıra hiperparametre optimizasyonu yoluyla QSVM modellerinin rafine edilmesi, hem doğruluk hem de hesaplama süresi açısından iyileştirmeler sağlayabilir. Ayrıca, kuantum sınıflandırma algoritmalarının farklı veri setleri ve daha karmaşık sınıflandırma görevleri üzerindeki performansını değerlendirmek, bu araştırmaların gerçek dünya uygulamaları için ne kadar uygun olduğunu anlamamıza yardımcı olacaktır.

**Çizelge-3: Quantum SVC 2. Kademe En İyi Yöntemde Grid Search Karşılaştırması**

| Model Numarası | Model | F1-Score | Çalışma Süresi (saniye) |
|---|---|---|---|
| 0 | reps=1, C=0.1, gamma=scale | 0.935936 | 34.665.970 |
| 1 | reps=1, C=0.1, gamma=auto | 0.957672 | 33.181.926 |
| 2 | reps=1, C=1, gamma=scale | 0.978120 | 31.604.035 |
| 3 | reps=1, C=1, gamma=auto | 0.978120 | 31.331.928 |
| **4** | **reps=1, C=10, gamma=scale** | **1.000.000** | **31.087.780** |
| 5 | reps=1, C=10, gamma=auto | 0.978645 | 34.686.012 |
| 6 | reps=2, C=0.1, gamma=scale | 0.854024 | 36.600.080 |
| 7 | reps=2, C=0.1, gamma=auto | 0.854024 | 37.226.757 |
| 8 | reps=2, C=1, gamma=scale | 0.955556 | 37.672.417 |
| 9 | reps=2, C=1, gamma=auto | 0.978120 | 36.095.115 |
| 10 | reps=2, C=10, gamma=scale | 0.978645 | 34.315.468 |
| 11 | reps=2, C=10, gamma=auto | 0.928313 | 35.639.712 |
| 12 | reps=3, C=0.1, gamma=scale | 0.540741 | 42.556.955 |
| 13 | reps=3, C=0.1, gamma=auto | 0.540741 | 42.073.531 |
| 14 | reps=3, C=1, gamma=scale | 0.797378 | 44.945.621 |
| 15 | reps=3, C=1, gamma=auto | 0.804495 | 46.811.185 |
| 16 | reps=3, C=10, gamma=scale | 0.843162 | 45.161.285 |
| 17 | reps=3, C=10, gamma=auto | 0.885470 | 45.514.616 |

Bu çalışma, kuantum makine öğreniminin olası avantajlarını ele almaya çalışmış, aynı zamanda pratik uygulamaların önündeki mevcut engelleri ve gelecekteki araştırma yönlerini de belirlemek için bu deneyi gerçekleştirmiştir. Gerçek kuantum bilgisayarlar üzerinde, daha büyük ölçekli veri setleri ile yapılan deneylerin tekrarı, kuantum makine öğreniminin ölçeklenebilirliği ve büyük veri problemleri karşısındaki etkinliği konusunda daha derin anlayışlar sağlayacaktır. Kuantum bilgi işlem teknolojisinin süregelen gelişimi, bu alanın zorluklarının üstesinden gelmeyi ve kuantum makine öğreniminin uygulama yelpazesini genişletmeyi vaat etmektedir. Bu araştırmanın reprodüktibilitesini sağlamak amacıyla, gerçekleştirilen deneylerin tüm metodolojileri, kodları ve veri setleri bir Google Colab defteri olarak erişime açılmıştır[29].

## Kaynakça